\documentclass[11pt, conference]{IEEEtran}
\usepackage{float}
\usepackage[dvipsnames]{xcolor}
\usepackage{tikz}
\usepackage{pgfplots}
\usetikzlibrary{patterns}
\usepackage{mathtools}
\usepackage{pst-node}
\usepackage{amsmath}
\usepackage{pdfpages}
\usetikzlibrary{positioning}

\usepackage{cite}
\usepackage{capt-of}
\date{August 9, 2016}
\usepackage{url}
\usepackage{comment}

\usepackage[
  pagebackref,
  bookmarksopen,
  bookmarksnumbered,
]{hyperref}

\hypersetup{
  pdfauthor={Toghiani-Rizi, Babak and Windmark, Marcus and Lind, Christofer and Svensson, Maria},
  % ...
}
\hypersetup{
    colorlinks=true,
    linkcolor=blue,
    citecolor={cyan!80!black},
    filecolor=magenta,      
    urlcolor=cyan,
}

\usepackage[nameinlink]{cleveref}
\usepackage{bookmark}

\def\figureWidth{0.4}

\pgfplotsset{compat=1.12}

\makeatletter
\def\ps@headings{%
  \def\@oddfoot{\scriptsize \@date\hfil \thepage}
  \def\@evenfoot{\scriptsize \@date\hfil \thepage}
}

\def\ps@IEEEtitlepagestyle{
  \def\@oddfoot{\scriptsize \@date\hfil \thepage}
  \def\@evenfoot{\scriptsize \@date\hfil \thepage}
}
\makeatother

\begin{document}
%
\title{Static Gesture Recognition \\ \huge{using Leap motion}}

\author{
  \IEEEauthorblockN{Babak Toghiani-Rizi}
  \IEEEauthorblockA{
    bato9963@student.uu.se \\
    Department of Information Technology\\
    Uppsala University
  }\\


  \IEEEauthorblockN{Christofer Lind}
  \IEEEauthorblockA{
    chli9685@student.uu.se\\
    Department of Information Technology\\
    Uppsala University
  }

  \and

  \IEEEauthorblockN{Maria Svensson}
  \IEEEauthorblockA{
    masv6227@student.uu.se\\
    Department of Information Technology\\
    Uppsala University
  }\\


  \IEEEauthorblockN{Marcus Windmark}
  \IEEEauthorblockA{
    mawi2661@student.uu.se\\
    Department of Information Technology\\
    Uppsala University
  }
}

\maketitle


\begin{abstract}
In this report, an automated bartender system was developed for making orders in a bar using hand gestures. The gesture recognition of the system was developed using Machine Learning techniques, where the model was trained to classify gestures using collected data. The final model used in the system reached an average accuracy of 95\%. The system raised ethical concerns both in terms of user interaction and having such a system in a real world scenario, but it could initially work as a complement to a real bartender.
\end{abstract}

\IEEEpeerreviewmaketitle

\section{Introduction} 
Any noisy environment can impose difficulties in using verbal communication, making it difficult to e.g. order drinks or food in a crowded bar. In cases like these, using hand gestures to order instead of talking is an alternative for a possibly better way of communicating. 

In this project, an automated bartender system was developed to ease similar situations in a bar. The system functioned by using a Leap Motion~\cite{weichert2013analysis} sensor to recognize hands, as well as Machine Learning techniques to train a model to classify hand gestures into actions. The system allowed the user to order any amount of drinks or food, undo actions and finally order using both cash and credit payment.

Accompanying the report is a blog~\cite{blog}, discussing the process of the group throughout the project.

\section{Feature processing} 
When choosing between the Leap Motion~\cite{weichert2013analysis} and Kinect~\cite{zhang2012microsoft}, we had to consider what valuable data each unit could extract that would make the process of constructing a feature vector as efficient as possible. Though we initially considered the Kinect, the final pick eventually fell on Leap Motion due to the fact that we wanted to use hand gestures. Kinect lacked any specific support for hand gestures, in comparison to Leap Motion that had an extensive API that enables extraction of a large range of different data regarding the hands.

For the feature vector, the data extracted included each finger's \textit{x} and \textit{y} coordinates (according to the grid that Leap Motion uses), extracted from both hands, as well as the center of both palms. The \textit{z}-coordinate, measuring the distance from the sensor, was in this case not saved, as the feature vector was supposed to be independent of this factor and solely rely on the relative positions of the fingers.

\begin{figure}[ht]
  \centering
  \includegraphics[scale=\figureWidth]{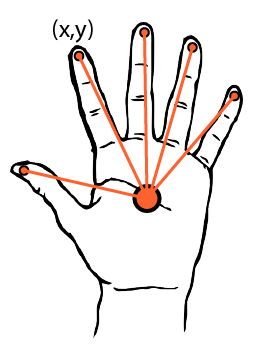}
  \caption{The data used in the feature vector was based on the distance between the palm center and the fingertips. }
  \label{fig:hand}
\end{figure}

\subsection{Pre-processing}
\label{sec:preprocessing}

Using the raw data extracted from the sensor, the feature first step of creating the feature vector was to make it with a length of 10, one for each finger using two hands. By calculating the length from each fingertip to the palm of the respective hand, a relative representation of each fingertip position was created, as seen in \cref{fig:hand}. To obtain these distances, the Euclidean distance in \cref{eq:eucl} was calculated, resulting in an array of 10 values.

\begin{equation}
d(p,q) = \sqrt[]{(q_{1}-p_{1})^{2} + (q_{1}-p_{1})^{2}}
\label{eq:eucl}
\end{equation}

As a method of abstraction, to study the relation between the fingers, regardless of the size of the hand, the data was normalized. However, since an important aspect was to conserve the co-relation between the fingers on each hand, all 10 distances were normalized in respect to their hand using the normalization formula in \cref{eq:norm}. The resulting normalization had values in the interval [0, 1].

\begin{equation}
x'_{h_{1},h_{2}}  =  \frac{x_{h}-\text{min}(x_{h})}{\text{max}(x_{h}) - \text{min}(x_{h})} \\
\label{eq:norm}
\end{equation}

\subsection{Dataset}
\begin{figure}[ht]
  \centering
  \includegraphics[scale=\figureWidth]{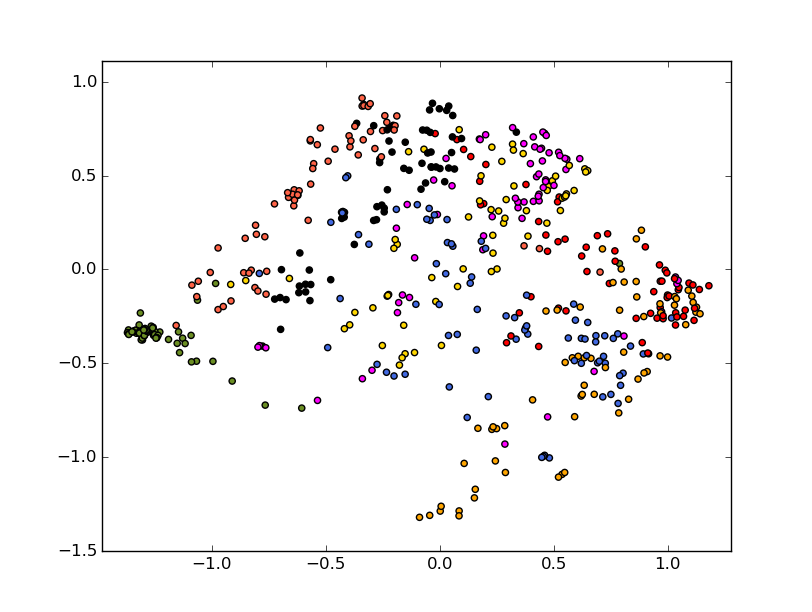}
  \caption{The 10-dimensional dataset was reduced to 2 dimensions with Principal Component Analysis, showing the eight different gesture-classes: Init in green, Alcohol in yellow, Non-alcohol in black, Food in pale red, Undo in orange, Checkout in red, Cash in magenta and Credit in blue.}
  \label{fig:dataset}
\end{figure}

Using an automated system of the pre-processing described in \cref{sec:preprocessing}, data samples were collected from roughly 20 participants. Each gesture was recorded approximately 3 times for each participant, resulting in a total amount of 528 samples. The distribution is shown in \cref{tab:dataset-samples}. Reducing the 10-dimensional feature vector to only 2 dimensions using Principal Component Analysis (PCA) resulted in \cref{fig:dataset}, where a bigger version can be seen in \cref{fig:dataset-appendix}.

\begin{table}[ht]
\centering
\begin{tabular}{llc}
 \multicolumn{1}{l}{\textbf{Index}} &\multicolumn{1}{l}{\textbf{Instruction}} & \multicolumn{1}{c}{\textbf{Samples}} \\
1 & Init 		& 	66 \\
2 & Alcohol 	& 	63 \\
3 & Non-alcohol & 	63 \\
4 & Food 		& 	65 \\
5 & Undo 		& 	64 \\
6 & Checkout 	& 	64 \\
7 & Cash 		& 	63 \\
8 & Credit 		& 	80 \\
\textbf{Total}	& & \textbf{528 samples}  	\\ \\
\end{tabular}
\caption{Distribution of the gesture samples in the dataset.}
\label{tab:dataset-samples}
\end{table}

\section{Training}
\label{sec:training}

In order to evaluate and optimize the performance of the system, three different machine learning techniques were used to train different models for classifying the gesture inputs. The model using k-Nearest Neighbor~\cite{peterson2009k} (kNN) was used as our main model in the final system, and two other types of Neural Network models, Multilayer Perceptron~\cite{neural-networks-foundation} (MLP) and Multinomial Logistic Regression\cite{starkweather2011multinomial} (MLR), were used as comparative models.

\subsection{k-Nearest Neighbor}
The main model used in the final system was k-Nearest Neighbor~\cite{peterson2009k}, a pattern recognition algorithm used for both classification as well as regression. We trained it by setting our \textit{k}, representing the number of nearest neighbors, to 2 neighbors.

\subsection{Multilayer Perceptron}
The first Neural Network model we trained was a TensorFlow\cite{abaditensorflow} based model, using Multilayer Perceptron~\cite{neural-networks-foundation}, with 10 input nodes (from the feature vector), 10 hidden nodes and 1 output node with eight different outputs, each representing one of the gestures. The model was trained using Gradient Descent~\cite{engelbrecht2007computational} as its learning heuristic and a step length of 0.01.

\subsection{Multinomial Logistic Regression}
The second Neural Network model trained was a Multinomial Logistic Regression (MLR)~\cite{starkweather2011multinomial}, also know as Softmax, which has the same base on TensorFlow as MLP. MLR is a classification method that generalizes the logistic regression to a multiclass problem by predicting probabilities of the different outcomes.

\section{Results and discussion} 
The dataset, containing eight classes with a total of 528 samples, as shown in \cref{fig:dataset}, gives an overview of how the feature vectors of the different gestures relate to each other. There are a few clusters that can be separated by only looking at the figure, such as the gestures for Init and Food, the classes are mostly overlapping. It is worth nothing that the 10 dimensions of the feature vector have been reduced by Principal Component Analysis to only 2 dimensions in this figure, so the distances that here seem to fully overlap has some distance. 

In this section, the results of the two approaches of classifying the eight gestures are presented and discussed. All results have been calculated using an average from testing the models multiple times. 

\subsection{Using k-Nearest Neighbor}
After having trained the kNN classifier, the performance of the model was tested using both Split Validation and Cross Validation. Both of these validation techniques were used to get a better overview of the general performance of the classifier.

The classification report received in one sample case of the kNN classifier is shown in \cref{fig:knn-classification-result} and the achieved results of both precision and recall were very good. Further analyzing these results shows that many of the classes got a perfect precision of 1.0, meaning that all samples in the predicted class belonged to the correct class. However, worth noticing is that the classes Cash and Food had a lower recall, which is a result of not all samples of the class being included in the predicted set. The confusion matrix in \cref{fig:knn-confusion-result} show another angle at the predicted samples of the model.

\begin{table}[ht]
\centering
\begin{tabular}{c | c c c c}
			& 	Precision	&	Recall	&  F1-score	&	Support \\  \hline
Alcohol   	&   1.00  		&   1.00	&  1.00   &     22 		\\
Cash   		&   1.00  		&   0.88  	&  0.94   &     17 \\
Checkout	&   0.78		&   0.95	&  0.86   &     19 \\
Credit   	&   1.00		&   1.00	&  1.00   &     24 \\
Food   		&   1.00		&   0.79	&  0.88   &     19 \\
Init   		&   1.00		&   1.00	&  1.00   &     15 \\
Non-Alcohol &   0.82		&   1.00	&  0.90   &     18 \\
Undo   		&   0.96		&   0.88	&  0.92   &     24 \\
\\
\textbf{Average / Total}    &   0.95   &   0.94   &   0.94    &   159 \\\\
\end{tabular}
\caption{A classification report of the knn classifier, using the 70/30 Split Validation method.}
\label{fig:knn-classification-result}
\end{table} 

\begin{table}[ht]
\centering
\begin{tabular}{c | c c c c c c c c}
Class	& 	1	&	2	&  3   &	4 	& 5 & 6 & 7 & 8 \\  \hline

1   	&   22  &   0	&  0   &    0	& 0 & 0 & 0 & 0	\\
2   	&   0  &   15   &  2   &    0 	& 0 & 0 & 0 & 0	\\
3		&   0	&   0	&  18  &    0 	& 0 & 0 & 0 & 1 \\
4   	&   0	&   0	&  0   &    24 	& 0 & 0 & 0 & 0 \\
5   	&   0	&   0	&  0   &    0 	& 15 & 0 & 4 & 0 \\
6   	&   0	&   0	&  0   &    0 	& 0 & 15 & 0 & 0 \\
7 		&   0	&   0	&  0   &    0 	& 0 & 0 & 18 & 0 \\
8   	&   0	&   0	&  3   &    0 	& 0 & 0 & 0 & 22 \\\\
\end{tabular}
\caption{Confusion matrix for knn using split validation.}
\label{fig:knn-confusion-result}
\end{table} 

\begin{tikzpicture}[
box/.style={draw,rectangle,minimum size=2cm,text width=1.5cm,align=left}]
\matrix (conmat) [row sep=.1cm,column sep=.1cm] {
\node (tpos) [box,
    label=left:\( \mathbf{p'} \),
    label=above:\( \mathbf{p} \),
    ] {True \\ positive};
&
\node (fneg) [box,
    label=above:\textbf{n},
    label=above right:\textbf{total},
    label=right:\( \mathrm{P}' \)] {False \\ negative};
\\
\node (fpos) [box,
    label=left:\( \mathbf{n'} \),
    label=below left:\textbf{total},
    label=below:P] {False \\ positive};
&
\node (tneg) [box,
    label=right:\( \mathrm{N}' \),
    label=below:N] {True \\ negative};
\\
};
\node [left=.05cm of conmat,text width=1.5cm,align=right] {\textbf{actual \\ value}};
\node [above=.05cm of conmat] {\textbf{prediction outcome}};
\end{tikzpicture}

\begin{figure}[ht]
  \centering
  \includegraphics[scale=\figureWidth]{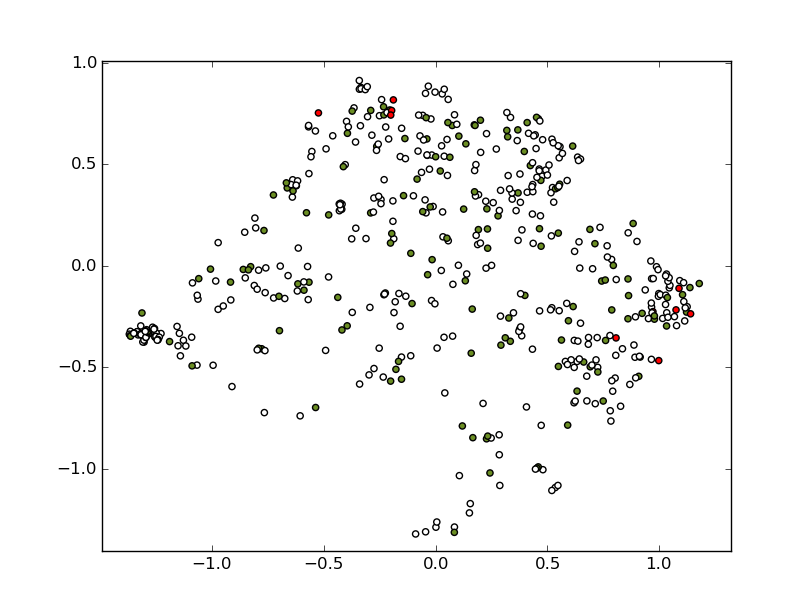}
  \caption{Misclassifications of the knn classifier using split validation. White dots represent training data, green dots correctly classified samples and red dots misclassifications.}
  \label{fig:knn-misclassifications}
\end{figure}

The plot of all mispredictions of the Split Validation is shown in \cref{fig:knn-misclassifications}, with a bigger version shown in \cref{fig:knn-misclassifications-appendix}, where the same dataset as in \cref{fig:dataset} was used. This figure shows the effect of the importance of which samples to use in training and testing. An example is the cluster of Food samples that all have been classified as Non-Alcohol, since the majority of the closest samples in this split happened to belong to that class. The simple algorithm kNN only took this into consideration and therefore made a misclassification.

The similarity between the feature vectors can be traced back to the definitions of the gestures. The difference between Food and Non-Alcohol is only the addition of the ring finger in the latter case. Undo and Checkout are two gestures differing only in which thumb was being extended and the results of the classifier confusing them is understandable.

The k-Fold cross validation technique was used with 5 folds to analyze the performance when using random partitions of the dataset in training and testing. The results are shown in \cref{fig:knn-kfold-result} with a total average accuracy of 84\%, but the results in the different partitions show yet again the variance of different splits.

\begin{table}[ht]
\centering
\begin{tabular}{c c c c c c}
k-Fold	& 	1	&	2	&  3   &	4 	& 5 \\  \hline
   	&   0.92 &  0.75 &  0.85 & 0.86 & 0.84	\\
\textbf{Average} & \textbf{0.84} \\ \\
\end{tabular}
\caption{The results of a k-fold cross validation of the knn model with a k of 5.}
\label{fig:knn-kfold-result}
\end{table} 





\subsection{Other Comparative Models}
One way of extending the evaluation of the performance of the main model, that used a kNN classifier, was to compare its performance to other classifier models, as explained in \cref{sec:training}, with a varying amount of training data. 

The evaluation was done by training each model with 50 random data samples, and then iteratively increase the amount of data by another 50 samples for each run and measure its average accuracy with each iteration. In the case of kNN, its accuracy average was measured using k-Fold Cross Validation (with \textit{k} set to 5), and in the case of the Neural Network-models, by taking the average of 5 runs.

\begin{figure}[ht]
\begin{tikzpicture}[scale=0.98]
\begin{axis}[
    title={Model Accuracy Comparison},
    xlabel={Number of Data Points},
    ylabel={Accuracy},
    xmin=50, xmax=500,
    ymin=60, ymax=100,
    xtick={0,100,200,300,400,500},
    scaled ticks=false,
    yticklabel=\pgfmathprintnumber{\tick}\,\%,
    legend pos=south east,
    ymajorgrids=true,
    grid style=dashed,
]
\addplot[
    color=blue,
    ]
    coordinates {
    (50,73.5)
    (100,77.1)
    (150,85.1)
    (200,88.2)
    (250,88.0)
    (300,92.0)
    (350,92.9)
    (400,95.1)
    (450,94.7)
    (500,94.8)
    };
  
\addplot[
    color=red,
    ]
    coordinates {
    (50,70.0)
    (100,72.0)
    (150,77.8)
    (200,81.3)
    (250,80.3)
    (300,79.8)
    (350,82.2)
    (400,86.3)
    (450,88.5)
    (500,86.6)
    };

\addplot[
    color=green,
    ]
    coordinates {
    (50,87.5)
    (100,90.3)
    (150,84.8)
    (200,83.6)
    (250,89.5)
    (300,87.9)
    (350,86.8)
    (400,85.8)
    (450,91.3)
    (500,91.3)
    };

    \legend{kNN, MLP, MLR, test}
\end{axis}
\end{tikzpicture}
\caption{Plotted average accuracy over number of data samples, showing how the models performed depending on the amount of data.}
\label{fig:acc-plots}
\end{figure}
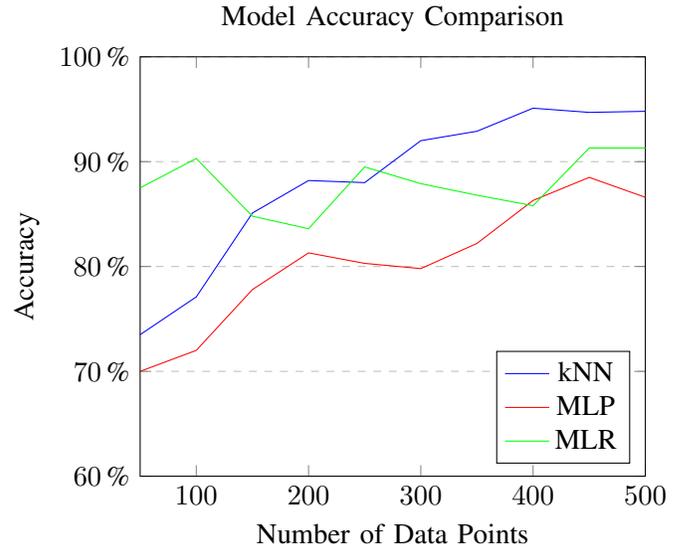

As seen in \cref{fig:acc-plots}, the plotted accuracy of the models over the amount of data they were trained on showed distinctive characteristics on sensitivity. The model using MLR displayed great resilience with minimal data, as it outperformed both of the other models when trained using only 50 data samples. However, as the numbers of data samples increased, the performance of the Neural Network models levelled and were eventually exceeded by the performance of the kNN-model. 

That kNN performed badly with low amounts of data is due to the definition of training that model, which simply consists of adding the training samples to the correct positions in the defined space. If the samples are few, it is unlikely that a test sample will be close to training samples of the same class.  

The results confirm that using the kNN model in the final system was appropriate, due to the large quantity of training data that was used.

\section{User Accessibility and Ethical Concerns}
Having an automated bartender system brings up several ethical concerns. In this section, a couple of them are discussed and suggested solutions to these problems are presented.

Communicating with hand gestures is often seen as the universal language, but an often overseen aspect is that hand gestures have different meaning in different cultures. An example is the thumbs up gesture, that is regarded as a positive sign in the Western world, but can be interpreted as a foul insult in Iraq.\cite{axtell1991gestures} When choosing the hand gestures for this system, the aim was to take these aspects into consideration. However, in the case of cultural differences, a solution could be to customize the system to use different hand gestures depending on the region.

An important part during the development of the system was to be agnostic to the type of user, to work well independently of the features of user's hands. One feature is the size and using the normalization technique, described in \cref{sec:preprocessing}, makes the system work with relative distances instead of absolute. The result is that the recognition works well with people of all hand sizes. Another aspect is the skin color, and by using the infrared lights in Leap Motion, instead of other methods such as tracking pixel colors, the system never considers this feature.

Both a visual interface as well as audio feedback was used to convey the interaction with the user in the bartender system. Reading the displayed text out loud allows users with visual impairment to use the system as well. The language is today limited to English, but extending it to more languages could be an improvement to include a larger audience.

However, the system does discriminate in the sense that you are required to have two hands with ten fingers in order to use it. It is also assumed that the co-relation of the lengths of your fingers can be generalized to fit within the dataset that the system was trained with. These are two major ethical concerns we were aware of, but chose not to address in order to limit the extent of the project.

Going beyond the technical approach to previously discussed ethical concerns, it is also important to discuss the extended responsibilities of a human bartender beyond serving and taking orders. Working in a bar demands an ability of making decisions regarding to who is appropriate to i.e. sell alcohol to. Selling alcohol to intoxicated individuals or minors may raise both moral and legal issues that the bartender has responsibility for.
Since the system is not a juridical entity, it cannot be held responsible for making such decisions. An automated bartender system requires this extra layer of decision-making which the system developed for this project does not support. To incorporate this, the system would become many times more complex, and even with full support, there would still need to be a human in charge to ensure that nothing goes wrong.

\section{Group evaluation} 
We started off this project by booking re-occurring meetings and workshops. By mostly working during these times, we made sure that everyone put a similar amount of effort into the project. Early on, a timeline, including a breakdown of the project into a number of tasks was made. Using the timeline as a base, no assignments of specific roles within the group were made, instead we simply divided the tasks depending on the current priority and progress. Work was often done in smaller groups of two and the groupings changed depending on the task. By utilizing the booked meetings and combining it with frequent communication, the group got off to a good start and continuously worked throughout the project. To conclude, all group members took responsibility and have contributed to the project.

\section{Future work} 
A potential future work would be to extend the hand gestures from the current use of static to continuous, which could make it more natural for human interaction. It would also extend the number of possible gestures that can be used.

Another part of the project that could be extended is the aspect of the gestures. The choice of the gestures could be a whole project on its own, with both a possible angle of studying the important relation between the gestures as well as making the gestures as user friendly as possible. Some of the gestures in this project were observed to frequently be misclassified and choosing better, more distinguishable, gestures could improve the accuracy.


\section{Conclusion}
The system worked successfully and using the classifying model of kNN, the results reached an accuracy of 95\%, which was shown to perform better than both implementations of neural networks with this amount of data.

It did however raise multiple ethical concerns, both in terms of using the system and the impact of having a similar system implemented in reality. Users are, for example, required to have two hands and ten fingers and the system currently does not have any limitation to who it could sell alcohol to.

The finished system would primarily only work as a good complement to a real life bartender, until many of these ethical issues have been addressed, but does in its current form not replace it completely.

The implementations used to achieve the results of this paper are available as open source code\footnote{https://github.com/windmark/static-gesture-recognition}.
\newpage

\bibliographystyle{IEEEtran}

\bibliography{bibliography}

\clearpage
\appendix

\renewcommand\thefigure{\thesection.\arabic{figure}}    
\setcounter{figure}{0}

\noindent
\begin{minipage}{\textwidth}
    \centering
	\includegraphics[width=1\textwidth]{images/dataset.png}
	\captionof{figure}{The 10-dimensional dataset was reduced to 2 dimensions with Principal Component Analysis, showing the eight different gesture-classes: Init in green, Alcohol in yellow, Non-alcohol in black, Food in pale red, Undo in orange, Checkout in red, Cash in magenta and Credit in blue.}
	\label{fig:dataset-appendix}
\end{minipage}

\clearpage
\noindent
\begin{minipage}{\textwidth}
    \centering
	\includegraphics[width=1\textwidth]{images/misspredictions-overview.png}
	\captionof{figure}{Misclassifications of the knn classifier using split validation. White dots represent training data, green dots correctly classified samples and red dots misclassifications.}
	\label{fig:knn-misclassifications-appendix}
\end{minipage}


\end{document}